\documentclass[5p, times]{elsarticle}
\makeatletter
\def\ps@pprintTitle{%
   \let\@oddhead\@empty
   \let\@evenhead\@empty
   \let\@oddfoot\@empty
   \let\@evenfoot\@oddfoot
}
\makeatother
\usepackage{bm}
\usepackage[svgnames,dvipsnames,table]{xcolor}
\usepackage[colorlinks]{hyperref}
\usepackage{amsmath}
\usepackage{amssymb}
\usepackage{adjustbox}
\usepackage{multirow}
\usepackage{tikz}
\usepackage{pgfplots}
\usepgfplotslibrary{patchplots}
\usetikzlibrary{trees,shapes,snakes,calc,matrix,positioning,fit,arrows,patterns,backgrounds,quotes,arrows.meta}
\usepackage{siunitx}
\usepackage{multicol}
\usepackage{caption}
\usepackage{subcaption} 
\usetikzlibrary{shapes.geometric}
\usetikzlibrary{shapes.arrows}
\usepgfplotslibrary{fillbetween}
\usepackage{array}
\usepackage{tabularx}
\usepackage{lipsum} 
 \usepackage{collcell}
\usepackage{hhline}
\usetikzlibrary{spy, calc}
\usepackage{array}
\usepackage{tabularx}
\usepackage{tabulary}
\usepackage{xspace}

\usepackage{bbding}
\usepackage{pifont}% http://ctan.org/pkg/pifont
\usepackage{diagbox}
\usepackage{booktabs}
\usepackage{makecell}
\usetikzlibrary{fit}
\usepackage{rotating}
\usepgfplotslibrary{colorbrewer}
\usepackage{fontawesome}
\usepackage{lscape}
\usepackage[para,online,flushleft]{threeparttable}
\usepackage{longtable}
\usepackage{tablefootnote}
\usepackage{makecell}
\usepackage{breakcites}
\usepackage{pifont}% http://ctan.org/pkg/pifont
\usepackage{hyphenat}
\begin{document}
%\verso{Given-name Surname \textit{et~al.}}
\begin{frontmatter}

\title{LE-CapsNet: A Light and Enhanced Capsule Network}
%\tnotetext[mytitlenote]{Fully documented templates are available in the elsarticle package on \href{http://www.ctan.org/tex-archive/macros/latex/contrib/elsarticle}{CTAN}.}

%% Group authors per affiliation:
%\author{Xin Yi\fnref{myfootnote}}
%\address{Radarweg 29, Amsterdam}
%\fntext[myfootnote]{Since 1880.}

\author[1]{Pouya Shiri}
% \cortext[cor1]{Corresponding author}
\ead{pouyashiri@uvic.ca}
\author[1]{Amirali Baniasadi}
\ead{amiralib@uvic.ca}
%\fntext[fn1]{This is author footnote for second author.}

%\author[2]{Given-name4 \snm{Surname4}}

\address[1]{Department of Electrical and Computer Engineering, University of Victoria, 3800 Finnerty Rd, Victoria, BC, V8P 5J2 Canada}

\begin{abstract}
Capsule Network (CapsNet) classifier has several advantages over CNNs, including better detection of images containing overlapping categories and higher accuracy on transformed images. Despite the advantages, CapsNet is slow due to its different structure. In addition, CapsNet is resource-hungry, includes many parameters and lags in accuracy compared to CNNs. In this work, we propose LE-CapsNet as a light, enhanced and more accurate variant of CapsNet. Using 3.8M weights, LE-CapsNet obtains 76.73\% accuracy on the CIFAR-10 dataset while performing inference 4x faster than CapsNet. In addition, our proposed network is more robust at detecting images with affine transformations compared to CapsNet. We achieve 94.37\% accuracy on the AffNIST dataset (compared to CapsNet's 90.52\%).
\end{abstract}


\end{frontmatter}

%\linenumbers
\section{Introduction}
Deep Convolutional Neural Networks (CNNs) have shown an excellent performance in classification competitions in Computer Vision and Image Processing \cite{Zhao2017a}. Despite its success, CNN has its own drawbacks. First, CNN tends to lose information wherever it uses pooling layers. Second, a CNN does not consider the relative spatial relationship between the low-level sub-objects that construct the main object in the classification task \cite{Xiang2018}. For this reason, CNN sometimes fails to classify images taken from a different viewpoint. Capsule Network (CapsNet) was introduced to address these drawbacks \cite{Shiri2021}.

\par
CapsNet considers the spatial relationship between objects constituting a category hierarchically and by creating a representation based on the part-to-whole relationships. The basic computational units in CapsNet are vectors (capsules). The size of these vectors (L2-Norm) represents the probability of the existence of a specific object. The direction of these vectors (the actual values in each dimension) contains the instantiation parameters of a specific object (e.g. pose information, thickness, etc). Due to its different structure and method of creating the representation, CapsNet classifies datasets with images containing overlapping categories more effectively. CapsNet is also better at detecting images with affine transformations. Accordingly, CapsNet captures the pose information and is, therefore, viewpoint-invariant \cite{Sabour2017}.

\par
Unfortunately, CapsNet is slow. This is mostly due to the employed Dynamic Routing (DR) algorithm. DR infers capsules of a layer based on the previous layer's capsules. CapsNet also uses a relatively high number of weights (parameters) compared to a conventional CNNs. Finally CapsNet falls behind conventional CNNs when it comes to accuracy \cite{shiri2020}. 

% In this work, we enhance CapsNet’s architecture and create a more powerful representation resulting in higher network accuracy. 

\par

% We introduce LE-CapsNet. LE-CapsNet improves performance by enhancing two different components: the feature extraction and the capsule formation. Building a better featuxtractor aims at providmore powerful capsules by extracting more useful features. Our improved capsule generation approach, aims at reducing the number of computations and network parameters. 

 \par

% CapsNet's feature extractor is very simple, consisting of only two consecutive convol,utional layers.

There have been several studies proposing alternative feature extractors to provide a better representation of the images. Some of these works are presented in section II. Most CapsNet variants, and the original version, create the first layer of capsules directly from the extracted features. This is done by reshaping the output of the feature extractor to capsules.

% Shiri et al \cite{Shiri2021} proposed CFC-CapsNet, which translates the extracted features to capsules using a new layer called Convolutional-Fully-Connected (CFC). This translation results in producing fewer yet stronger capsules and achieves higher classification accuracy. 

\par
In this work, we propose LE-CapsNet as a light and enhanced variant of CapsNet. We introduce the Primary Capsule Generation (PCG) module, that is responsible for extracting the primary capsules from the input image. This module includes a different feature extraction method. In our approach the features are extracted in different scales. Here the term scale refers to the depth of the feature extractor at each level, which is equal to its number of its convolutional layers. The multi-scale feature extraction provides a various set of features that all together constitute an effective representation of the image. This representation includes a mixture of low-level and high-level features, which in turn produces capsules providing a good representation of the input image. We present more details about this architecture in section IV. In addition, LE-CapsNet customizes several instances of the Convolutional Fully-Connected (CFC) \cite{Shiri2021} as a summarizing approach where the extracted features are transformed into a small number of capsules. Moreover, we use an enhanced decoder instead of the simple decoder used in conventional CapsNet. This alternative decoder generates a better reconstruction using fewer parameters. Finally, we use a modified capsule dropout method to facilitate network generalization. In summary, we make the following contributions.

% \par

% TTo summarize the extracted features, we customize several instances of the Convolutional Fully-Connected (CFC).

% We integrate the CFC layer \cite{Shiri2021} into the multi-scale feature extractor. We refer to this network LE-CapsNet (Light and Enhanced CapsNet). Here the CFC layer is applied at the end of each scale of the feature extractor, and the outputs of different scales are concatenated. In summary we make the following contributions. 

\begin{itemize}
    \item We introduce a faster alternative to CapsNet. Our solution is faster as it contains significantly fewer capsules. As a result, the capsules of the different layers are inferred faster and the capsule inference algorithm (Dynamic Routing algorithm) takes less time to finish. LE-CapsNet performs training and inference 2.4x and 4x faster.
    \item We achieve a higher classification accuracy. This is due to the integration of the PCG module and the alternative decoder that we use in this work.
    % This is due to the integration of a translator layer (CFC layer) and extracting features in different scales.
    LE-CapsNet creates a good representation of the input image. For example, on the CIFAR-10 dataset we improve accuracy from 71.69\% for CapsNet to 76.73\%. We compare some of the state-of-the-art works in section V.
    
    \item We reduce the number of parameters significantly. The PCG module builds on a novel method for producing significantly fewer number of capsules. In addition, the decoder used in LE-CapsNet includes a fewer number of parameters compared to the conventional decoder of CapsNe. LE-CapsNet includes 3.8M parameters (versus 11.7M of CapsNet) on the CIFAR-10 dataset.
    % This is mostly due to the usage of CFC layer, as there are fewer capsules generated at each scale. In addition, the multi-scale feature extractor includes fewer parameters compared to the original feature extractor. 

    \item We obtain a higher accuracy in robustness to affine transformations. LE-CapsNet improves the accuracy by 3.85\% reaching 94.37\%. The PCG module generates capsules out of extracted features in multiple scales. This advantage combined with the use of capsule translation module results in a better part-to-whole representation.
    
    % As a result, we improve the network's robustness to affine transformations applied to the input images.

\end{itemize}

This paper is organized as follows. We review related works in section II. Section III explains the background. We present LE-CapsNet in section IV. We present the experiments and their corresponding results in section V. We offer concluding remarks in section VI.

\section{Related Works}
Several works have focused on improving CapsNet. Some studies either improve the encoder or the decoder, while others propose a different inference algorithm and manipulate capsules differently. Rosario et al. \cite{DoRosario2019} propose Multi-Lane CapsNet (MLCN). MLCN achieves a higher accuracy using the same number of parameters as CapsNet. MLCN is also faster than CapsNet in training and inference and considers different lanes for each dimension of the digit capsules. 
% Lanes can have different types based on the number of convolutional layers used as well as the number of filters per convolutional layer. MLCN produces a more powerful representation of the data as it applies separate lanes for each dimension of the digit capsules. MLCN also applies a lane-dropout method that randomly dismisses 10\% of lanes. This is done to make the classification independent of specific lanes and to regularize the network. 
\par
\begin{figure*}%[H]
    \centering
    % \captionsetup{width=\textwidth,height=\textheight}
    \includegraphics[scale=0.7,keepaspectratio]{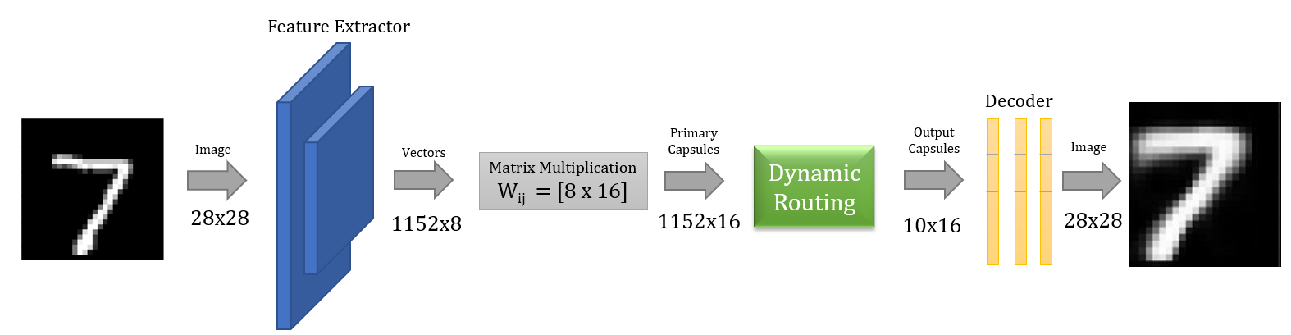}
    %[width=\textwidth,height=\textheight]
    \caption{The CapsNet's architecture. Vectors are formed by reshaping the extracted features from the feature extractor, then multiplied with a matrix to create PCs. The DR algorithm infers the output capsules by which the classification is performed. The decoder reconstructs the input image.}
    \label{fig:caps-arch}
\end{figure*}
Shiri et al. \cite{shiri2020} propose Quick-CapsNet (QCN). QCN alters the feature extractor of CapsNet and integrates a Fully-Connected layer. This results in producing a significantly fewer number of capsules which in turn accelerates the network. This network also alters and uses the class-independent decoder proposed by Rajasegaran et al \cite{Rajasegaran2019}. 
\par
Shiri et al. \cite{Shiri2021} propose Convolutional Fully-Connected CapsNet (CFC-CapsNet). They propose a new layer referred to as the CFC layer that takes care of translating the extracted features to the first layer of capsules. This translation summarizes the output of the feature extractor to a fewer number of capsules. The translation is learned during the training process. CFC-CapsNet achieves  higher accuracy compared to conventional CapsNet. The reduction of the number of capsules also results in a significant reduction in the number of parameters. 
\par
% Rajasegaran et al. \cite{Rajasegaran2019} propose DeepCaps as a deeper variant of CapsNet. DeepCaps has a deep feature extractor that includes several skip-connected blocks. This variant proposes a new decoder referred to as the class-independent decoder. This decoder uses deconvolutional layers instead of the Fully-Connected (FC) layers which was originally used in the CapsNet decoder.
\par
 Xiang et al \cite{Xiang2018} introduce Multi-Scale CapsNet (MS-CapsNet). MS-CapsNet extracts features in different scale. Moreover, the output of each scale is reshaped into capsules. This work considers capsules of different dimensionalities for each scale. After the predictions are made, they are concatenated to form the primary capsules. MS-CapsNet further implements a capsule dropout method making the network independent of specific capsules. 
%  This works as a regularization technique for the training process of the network. 

% XXX dacapsnet commented out
% \par
% Huang et al. propose a Dual Attention mechanism Capsule Network (DA-CapsNet) \cite{Huang2020}. This network integrates the attention mechanism into CapsNet after the convolution layers. The attention modules are also used after the primary capsules. The attention mechanism is used to scale data based on importance and hence results in more representative features and capsules. 
% This network obtains a high accuracy on small-scale datasets and performs a more accurate image reconstruction compared to CapsNet.

\par

\par
LE-CapsNet is designed to create an accurate representation of the image using significantly fewer number of capsules. As a result, unlike other related works,  this networks takes a more holistic approach aiming at parameter reduction, training and testing speedup and higher classification accuracy simultaneously. This is realized by introducing the PCG module and the improved reconstruction network (decoder). The PCG module is carefully designed to generate very few capsules with a high representativity of the input image.

\begin{figure}[htp]
    \centering
    % \captionsetup{width=\textwidth,height=\textheight}
    \includegraphics[width=0.5\textwidth,keepaspectratio, scale=0.7]{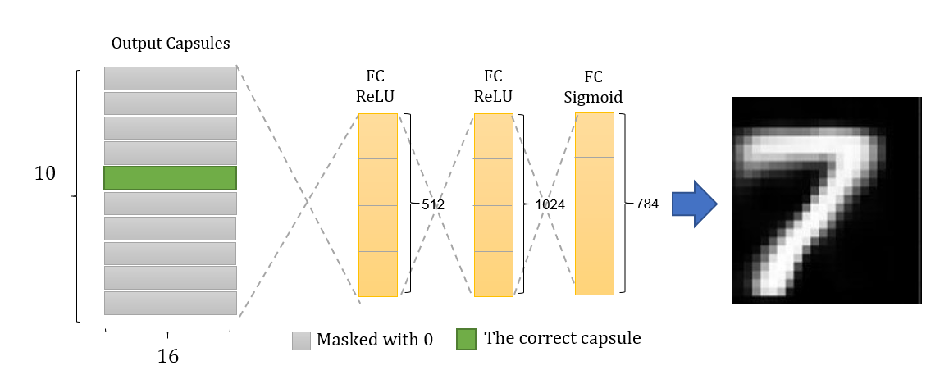}
    %[width=\textwidth,height=\textheight]
    \caption{CapsNet's Decoder. Consecutive FC layers are used to reconstruct the input image. All output capsules except for the correct one are masked out with zeroes.} 
    \label{fig:caps_dec}
\end{figure}

\section{Background}
\subsection{Capsule Network}
% Capsule network uses a basic computational unit different from that used by CNNs. This unit is a vector of neurons and is referred to as a capsule.

The architecture of CapsNet is shown in figure \ref{fig:caps-arch}. As the figure shows, there is a low-level feature extractor consisting of two convolutional layers and  its output is reshaped to vectors. The resulting vectors are then multiplied with a matrix. This is done to encode the spatial relationship between the vectors formed by the feature extractor. The resulting vectors are referred to as  Primary Capsules (PCs).
\par

The next stage in the architecture of CapsNet includes summarizing the PCs into the output capsules by which the classification is performed. This is done using Dynamic Routing (DR). In this stage all PCs contribute to each output capsule. The amount of each contribution is decided by coefficients that are determined by the DR algorithm. These coefficients are not learned during the training process. Instead, the DR algorithm determines them using an iterative process which relies on detecting agreements among PCs on activating output capsules.

\par
The number of output capsules equals the number of classes in the classification task. The output capsules are used to perform the classification: the capsule with the highest length corresponds to the correct category. The output capsules also hold additional information. Each dimension of an output capsule corresponds to a different instantiation property, e.g. deformation and pose. This is verified using the decoder in the CapsNet\cite{Sabour2017}. 
\par

CapsNet includes a decoder network that reconstructs the input image using the output capsules. The decoder network is shown in figure \ref{fig:caps_dec}. This decoder consists of consecutive Fully-Connected (FC) layers. Comparing the resulting images to the input images is used to create an additional term in the network loss function: the reconstruction loss. This is used as a regularizer for the training process.  

\par
 The loss function of CapsNet includes another loss term referred to as the margin loss. All the capsules contribute to this loss term based on the following equation:
 \begin{equation}
      L_k = T_k \max(0,m^+-||V_k||)^2 + \lambda(1-T_k)\max(0,||V_k||-m^-)^2 
     \label{eq_loss}
 \end{equation}

where $||V_k||$ is the length of the k-th output capsule, $T_k$ is 0 or 1 based on whether the prediction of the k-th output capsule is correct or not, $\lambda$ is the coefficient considered for wrong predictions and $m^+$ and $m^-$ are used to prevent the capsules with a very low (or very high) probability from affecting the margin loss.

\subsection{Class-Independent Decoder}
The class-independent decoder was introduced by Rajasegaran et al. \cite{Rajasegaran}. This decoder consists of deconvolutional layers instead of FC layers. In addition, the output capsules are fed to the decoder in a different manner. Employing this decoder results in better reconstructions as deconvolutional layers capture spatial relationships better than FC Layers. Also, due to the weight-sharing property in deconvolutional layers, these layers include fewer parameters compared to FC layers. 

\par

\begin{figure}[htp]
    \centering
    % \captionsetup{width=\textwidth,height=\textheight}
    \includegraphics[width=0.5\textwidth,keepaspectratio]{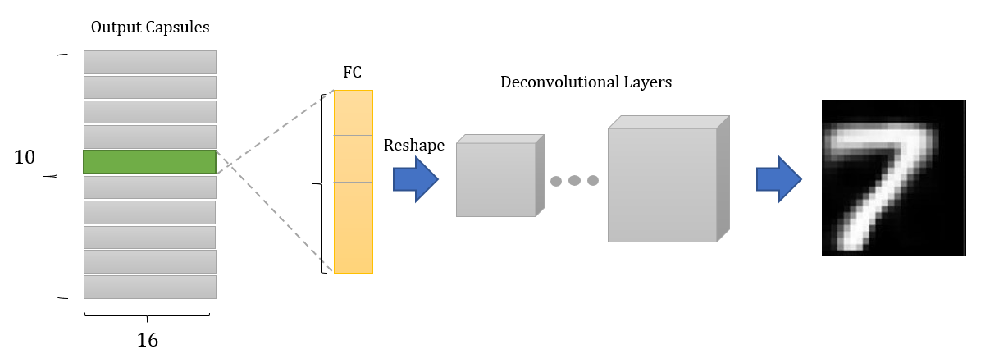}
    %[width=\textwidth,height=\textheight]
    \caption{Class-Independent Decoder for CapsNet}
    \label{fig:caps-newdec}
\end{figure}

As shown in figure \ref{fig:caps_dec}, the decoder originally used in CapsNet includes all output capsules, but masks all but the correct capsule with zeros. The correct prediction is based on the ground-truth during the training. During inference, the length of the capsule makes the final decision: the capsule with the highest length is the correct capsule. This decoder is considered as class-independent Since all categories are present (though sometimes masked) as the input level. 

\par
We show the architecture of the class-independent decoder in \ref{fig:caps-newdec}. This decoder consists of consecutive deconvolutional layers which progressively enlarge the representation and finally reconstruct the input image. This decoder only keeps the correct capsule and instead of masking the rest,  disregards them. The decoder is class-independent as all categories are treated equally.

\begin{figure*}[htp]
    \centering
    % \captionsetup{width=\textwidth,height=\textheight}
    \includegraphics[keepaspectratio,width=\linewidth]{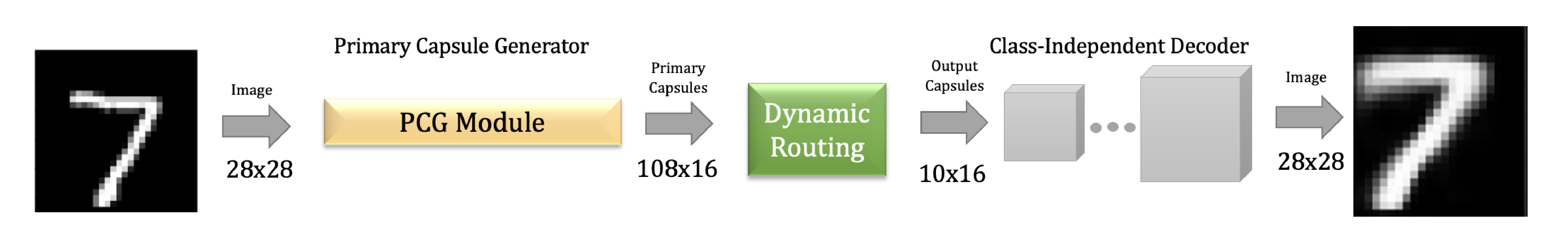}
    %[width=\textwidth,height=\textheight]
    \caption{LE-CapsNet architecture. The network uses a PC Generator module to generate capsules out of the input image. This module produces 90\% fewer capsules (108 vs. 1152) on the MNIST dataset. This reduction leads to a faster and lighter (in terms of number of parameters) network.}
    \label{fig:LE-arch}
\end{figure*}

% \subsection{Multi-Scale Capsule Network}

% LE-CapsNet extracts features in multiple scales. These scales differ in depth. The number of convolutional layers used in each scale determines how deep each scale is. Deeper scales are used to create representations of features of a higher level, whereas shallow scales correspond to low-level features. This helps in creating a better representation of the input images.

\subsection{Convolutional Fully Connected (CFC) Layer}
\begin{figure}[htp]
    \centering
    % \captionsetup{width=\textwidth,height=\textheight}
    \includegraphics[keepaspectratio,scale=0.5]{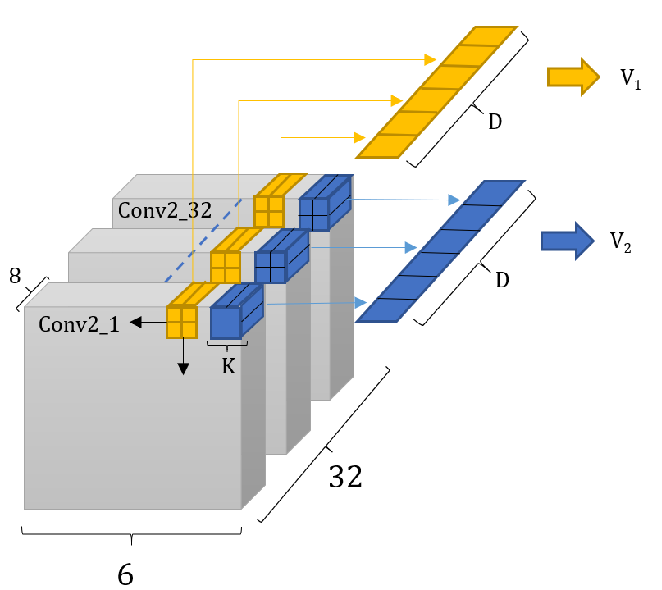}
    %[width=\textwidth,height=\textheight]
    \caption{The CFC layer. For each stride, vectors are formed from spatially correlated regions of the output feature map.}
    \label{fig:cfc_layer}
\end{figure}

% We customize and employ several instances of the Convolutional Fully-Connected (CFC) layer to generate the capsules using the extracted features.
We present this layer in figure \ref{fig:cfc_layer}. The CFC layer \cite{Shiri2021} is designed to translate the low-level extracted features into capsules. The CFC layer works like a convolutional layer. The only difference is that the CFC layer uses multiple kernels, whereas the convolutional layer convolves a single kernel with the input layer. In the CFC layer, each kernel is responsible for translating a specific spatial location of the extracted features to capsules. CapsNet works by creating a part-to-whole relationship between the low-level sub-objects that constitute the main object. The way the CFC layer summarizes the extracted features creates a robust and powerful starting point for forming this part-to-whole relationship. 
\par
%\textbf{The CFC layer has two parameters: the kernel size (K) and the output dimensionality for the created capsules (D). Choosing a higher K for the CFC layer results in fewer numbers of capsules but results in a slightly lower classification accuracy\cite{Shiri2021}. Different values for D also affect different aspects of the network, but the change is marginal.
%WE CAN REMOVE THIS SECTION (SPECS OF CFC LAYER). WHAT DO YOU THINK?} 

\section{LE-CapsNet}
\begin{figure*}[htp]
    \centering
    % \captionsetup{width=\textwidth,height=\textheight}
    \includegraphics[width=\textwidth,keepaspectratio]{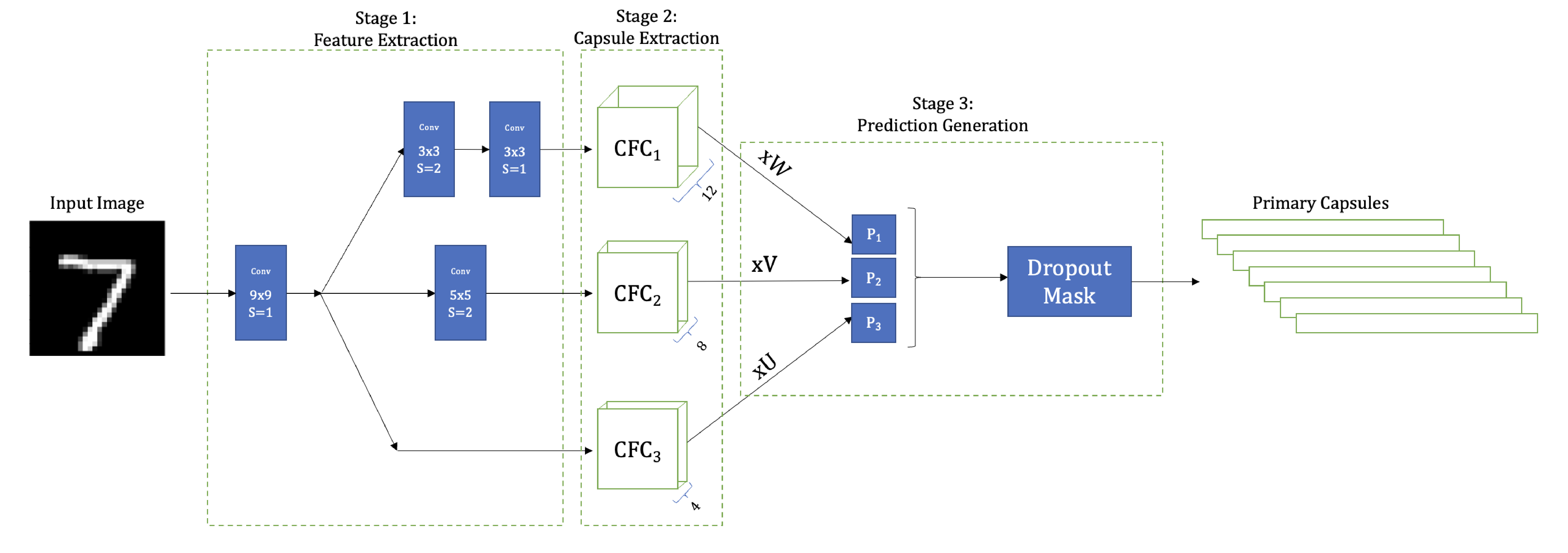}
    %[width=\textwidth,height=\textheight]
    \caption{The architecture of the PCG module. This module consists of three steps. First, features are extracted in three scales. The second step consists of using CFC layers on top of each scale. After translating the extracted features to capsules of different lengths, capsules are multiplied with different matrices in the third step and the result is concatenated. Capsule dropout mechanism to regularize training.}
    \label{fig:PCG-arch}
\end{figure*}

In figure \ref{fig:LE-arch} we present the architecture of LE-CapsNet. LE-CapsNet uses a Primary Capsule Generator (PCG) module to generate the primary capsules from the input image. As the figure shows, there is a significant reduction in the number of produced capsules using this module. For the MNIST dataset, the number of PCs is reduced from 1152 to 108 capsules \footnote{The choice of dataset impacts these numbers as feature extraction results in extracting the different amount of information (a different feature map size) for different datasets and based on the input image size.}. Figure \ref{fig:PCG-arch} shows the architecture of the PCG module. This module consists of three steps. The first step is the low-level feature extraction step. In this step, we use a multi-scale sub-network instead of the two convolutional layers used in CapsNet. This sub-network consists of three scales, each with a different depth. The depth of each scale corresponds to the number of convolutional layers used in that scale. Deeper scales are used to create representations of features of a higher level, whereas shallow scales correspond to low-level features. This helps in creating a better representation of the input images.
% In this work, we replace the basic feature extractor of CapsNet with a multi-scale feature extractor. This is done to create a more powerful representation from which capsules are created. Afterward, we use different CFC layers for each of the scales in the feature extractor. Figure \ref{fig:LE-arch} shows the proposed architecture of LE-CapsNet. 
\par 
The second step in the PCG module uses multiple CFC layers (one for each scale) to translate the extracted features to vectors. We modify CFC layers by varying the dimensionality of the output based on how deep a specific scale in the feature extraction is. For scales that extract features deeper, the output dimensionality of the CFC layer is chosen to be higher. This provides more outputs and hence better representation. In order to keep important information while summarizing the extracted information, we use more output neurons for deeper scales. For more shallow scales, we create a smaller summary of information, and the output dimensionality is smaller.
\par
The third step of the PCG module includes the affine matrix transformation and the capsule dropout. The output vectors of the second step are multiplied with the weight matrix (the affine transform matrix multiplication stage). The multiplication is done individually for each scale to encode the spatial relationship between the vectors of each scale separately. Contrary to CapsNet, the generated primary capsules are a mixture of translated low-level, medium-level and high-level features. As such, we expect to create a better representation leading to higher test accuracy. Note that there are three different sets of primary capsules generated by this method. The capsule dropout is explained in the next section in detail.
\par
LE-CapsNet also implements the class-independent decoder. This has two reasons: reducing the number of parameters and providing a more accurate reconstruction. This, in turn, helps create a more powerful network in terms of classification accuracy and generalization ability.

\subsection{Capsule Dropout}
The dropout method is used frequently to remove neurons randomly during training, to make the network more robust and not dependent on specific neurons. Dropout improves  network generalization through regularizing the training process. This general approach does not work well with CapsNet, since dropping neurons changes the direction of capsules and could affect the representation negatively\cite{Xiang2018}. Xiang et al. \cite{Xiang2018} introduced an alternative dropout method  in which an entire capsule is randomly dropped during training. We use this dropout method in LE-CapsNet.

% Removing neurons inside a capsule results in an incomplete capsule as it changes its direction and length.

% The conventional dropout method (randomly dropping neurons) does not work well with CapsNet as dropping neurons change the direction of capsules and affects the representation. In this paper, we use the same dropout method.

% [---WE need a bit more on Xiang's dropout method. Basically, we are saying very little. add more details early in the paragraph-----]

\section{Experiments and Results}
This section explains the experiments we performed and presents the results in detail. 
\subsection{Datasets}
In general, CapsNet does not perform well on large-scale datasets with a high number of categories. We test LE-CapsNet on certain target datasets, i.e., Fashion-MNIST (F-MNIST) \cite{Xiao2017}, SVHN \cite{Netzer2011} and CIFAR-10 \cite{Krizhevsky2009} datasets. Table \ref{table:table_dsets} lists the datasets used for testing the performance of our proposed network. We further test the network against the AffNIST dataset to verify the robustness of applying affine transformations to the input images.

% copied
\begin{table}[htp]
\caption{A List of the datasets used for testing the performance of LE-CapsNet. } % title of Table
\centering % used for centering table
\resizebox{\linewidth}{!}{\begin{tabular}[width=\linewidth]{|c|c| c| c| c|c|} % centered columns (4 columns)
\hline \hline
\textbf{Name} & \textbf{Image Size} & \textbf{\#Channels} & \textbf{Training samples} & \textbf{Test Samples}  \\ [1ex]

\hline
MNIST & 28x28 & 1 & 50,000 & 10,000  \\ [1ex]
\hline
F-MNIST & 28x28 & 1 & 50,000 & 10,000 \\ [1ex]
\hline
SVHN & 32x32 & 3 & 73,257 & 26,032 \\ [1ex]
\hline
CIFAR-10 & 32x32 & 3 & 50,000 & 10,000  \\ [1ex]

% adds vertical space
\hline %inserts a single line
\end{tabular}}
\label{table:table_dsets} % is used to refer this table in the text
\end{table}

% Table \ref{table:table_dsets} lists the datasets used for testing the performance of our proposed network. Fashion-MNIST (F-MNIST) is a dataset containing 28x28 grey-scale images of 10 different types of clothing. This dataset follows the same format as the MNIST dataset \cite{LECUN}. However, the images are more complex in the F-MNIST dataset. There are 50,000 and 10,000 images in the training and testing sets of this dataset respectively. 

% \par
% Cifar-10 and SVHN datasets both consist of 32x32 colored images and contain 10 categories. The cifar-10 dataset includes images of different objects and animals e.g. frogs, and planes. SVHN dataset includes cropped images of the different digits in house numbers. There are 50,000 and 10,000 images in the training and testing sets of the Cifar-10 dataset, while there are 73,257 and 26,032 images in the training and testing sets of the SVHN dataset. The cifar-10 dataset is the most complex dataset among the three datasets we used in this research. This is due to the various background used in the images. 

\subsection{Experiments Settings}
We implement LE-CapsNet on top of the PyTorch implementation of CapsNet \footnote{https://github.com/gram-ai/capsule-networks}. We use a 2080Ti GPU with 11GB VRAM for running the experiments. Similar to CFC-CapsNet, we perform hard training for all experiments. Using this method, we train the network twice. The second round of training uses a loss function with tighter bounds for the loss of capsule predictions (different values for m+ and m- values in equation \ref{eq_loss}). All the experiments are repeated five times. As results show very few variations, we report the average values. We used the Adam optimizer and the default learning rate of LR=0.001. In addition, we apply an exponential decay of gamma=0.96 to the learning rate. We set the batch size to 128.

\subsection{Network Accuracy}
Table \ref{table:tres_acc} reports network testing accuracy \footnote{Note that in the interest of a fair comparison, we implemented LE-CapsNet in the PyTorch version of CapsNet. CapsNet and CFC-CapsNet are implemented using the same version. In addition, we implemented and tested MS-CapsNet in the same framework.}. 
% XXX talk about having access to the implementation of some works!
% fix the decoder of LE-CapsNet in the main image!
With 75.75\%, 93.02\% and 93.14\%, LE-CapsNet achieves a competitive accuracy on CIFAR-10, SVHN and F-MNIST datasets. This is mostly due to the PCG module's integration, as it effectively produces capsules representing the input image. In addition, the class-independent decoder provides a better reconstruction of the images and helps the network generalize even more. DA-CapsNet obtains a higher accuracy for all three datasets. However, the number of parameters used in this architecture is unknown, making a fair comparison impossible. Still, we report results for this architecture in the interest of full disclosure. 

\par
 For the dropout method, we tested different dropout rates of 5\%, 10\% and 40\% for LE-CapsNet and reported the best result. The dropout improved test accuracy only for the CIFAR-10 dataset (75.75\% went up to 76.73\%). For the other two datasets, dropout resulted in a slight drop in accuracy. Since the images are small in size, there are few capsules and dropping capsules could sometimes result in the loss of important information. 

\newcommand{\specialcell}[2][c]{%
  \begin{tabular}[#1]{@{}c@{}}#2\end{tabular}}
  
\begin{table}[htp]
\caption{The network testing accuracy for different architectures} % title of Table
\centering % used for centering table
%\resizebox{\linewidth}{!}
%\begin{tabular*}[width=\linewidth]{|c|c| c| c| c|} % centered 
%columns (4 columns)
\begin{tabular}[width=\linewidth]{|c|c| c| c|}
\hline\hline %inserts double horizontal lines
\textbf{Architecture} & \textbf{\specialcell{Accuracy \\ (CIFAR-10)}} & \textbf{\specialcell{Accuracy \\ (SVHN)}} & \textbf{\specialcell{Accuracy \\ (F-MNIST)}} \\ [0.5ex] % inserts table
%heading
\hline % inserts a single horizontal line

CapsNet \cite{Sabour2017} & 71.69\% & 92.70\% & 91.37\% \\ [1ex] % inserting body of the table
\hline
CFC-CapsNet \cite{Shiri2021} & 73.15\% & 93.29\% & 92.86\% \\ [1ex]

\hline
MS-CapsNet \cite{Xiang2018} & 72.30\% & 92.68\% & 92.26\% \\ [1ex]

\hline
MLCN \cite{DoRosario2019} & 75.18\% & - & 92.63\%  \\ [1ex]

% \hline 
% HitNet \cite{Deli2018} & 73.30\% & 94.50\% & 92.30\% \\ [1ex]

% \hline
% DA-CapsNet \cite{Huang2020} & 85.47\% & 94.82\% & 93.98\% \\ [1ex]

\hline
LE-CapsNet & 75.75\% & 93.02\% & 93.14\% \\ [1ex]

\hline
\textbf{\specialcell{LE-CapsNet \\ (+ Dropout)}} & 76.73\% & 92.62\% & 93.04\% \\ [1ex]

% adds vertical space
\hline %inserts a single line
\end{tabular}
\label{table:tres_acc} % is used to refer this table in the text
\end{table}
% As for DA-CapsNet, the authors have not disclosed the number of parameters.
\subsection{Number of Parameters}
Table \ref{table:tres_param} reports the number of parameters \footnote{For the MLCN architecture, we are reporting the number of parameters for the variant with the highest accuracy. }. With only 3.8M weights for the CIFAR-10 dataset, LE-CapsNet employs the fewest number of parameters.
This is due to the integration of the PCG module (producing fewer capsules) and the class-independent decoder (deconvolutional reconstruction)\footnote{Note that MS-CapsNet uses 13x13 kernels for the first convolutional layer. We tested this network using the PyTorch implementation, resulting in very low accuracy. Therefore we also decided to use the kernel size used in other datasets (9x9 kernel) for CIFAR-10. This explains why we require a high number of parameters (in contrast to what authors state \cite{Xiang2018}) for CIFAR-10.}.
% The multi-scale feature extractor includes fewer weights compared to CapsNet's feature extractor. In addition, the CFC layer summarizes the extracted features to fewer capsules compared to CapsNet. Therefore, we achieve the benefits of both methods in LE-CapsNet 

\begin{table}[htp]
\caption{The number of parameters for different architectures. LE-CapsNet is the lightest network. } % title of Table
\centering % used for centering table
%\resizebox{\linewidth}{!}
%\begin{tabular*}[width=\linewidth]{|c|c| c| c| c|} % centered 
%columns (4 columns)
\begin{tabular}[width=\linewidth]{|c|c| c|}
\hline\hline %inserts double horizontal lines
\textbf{Architecture} & \textbf{\specialcell{Parameters \\ (CIFAR-10)}} &  \textbf{\specialcell{Parameters \\ (F-MNIST)}} \\ [0.5ex] % inserts table
%heading
\hline % inserts a single horizontal line

CapsNet \cite{Sabour2017} & 11.7M & 8.2M  \\ [1ex] % inserting body of the table
\hline
CFC-CapsNet \cite{Shiri2021} & 5.9M & 5.7M  \\ [1ex]

\hline
MS-CapsNet \cite{Xiang2018} & 13.9M & 8.3M \\ [1ex]

\hline
MLCN \cite{DoRosario2019} & 14.2M & 10.6M \\ [1ex]

% \hline
% HitNet \cite{Deli2018} & 8.9M & 8.9M \\ [1ex]

% \hline
% DA-CapsNet \cite{Huang2020} & - & - \\[1ex]
\hline
LE-CapsNet & 3.8M & 3.4M  \\ [1ex]

% adds vertical space
\hline %inserts a single line
\end{tabular}
\label{table:tres_param} % is used to refer this table in the text
\end{table}

\subsection{Network Training and Inference Time}

\begin{figure}[htp]
    \centering
    % \captionsetup{width=\textwidth,height=\textheight}
    \includegraphics[keepaspectratio, scale=0.2]{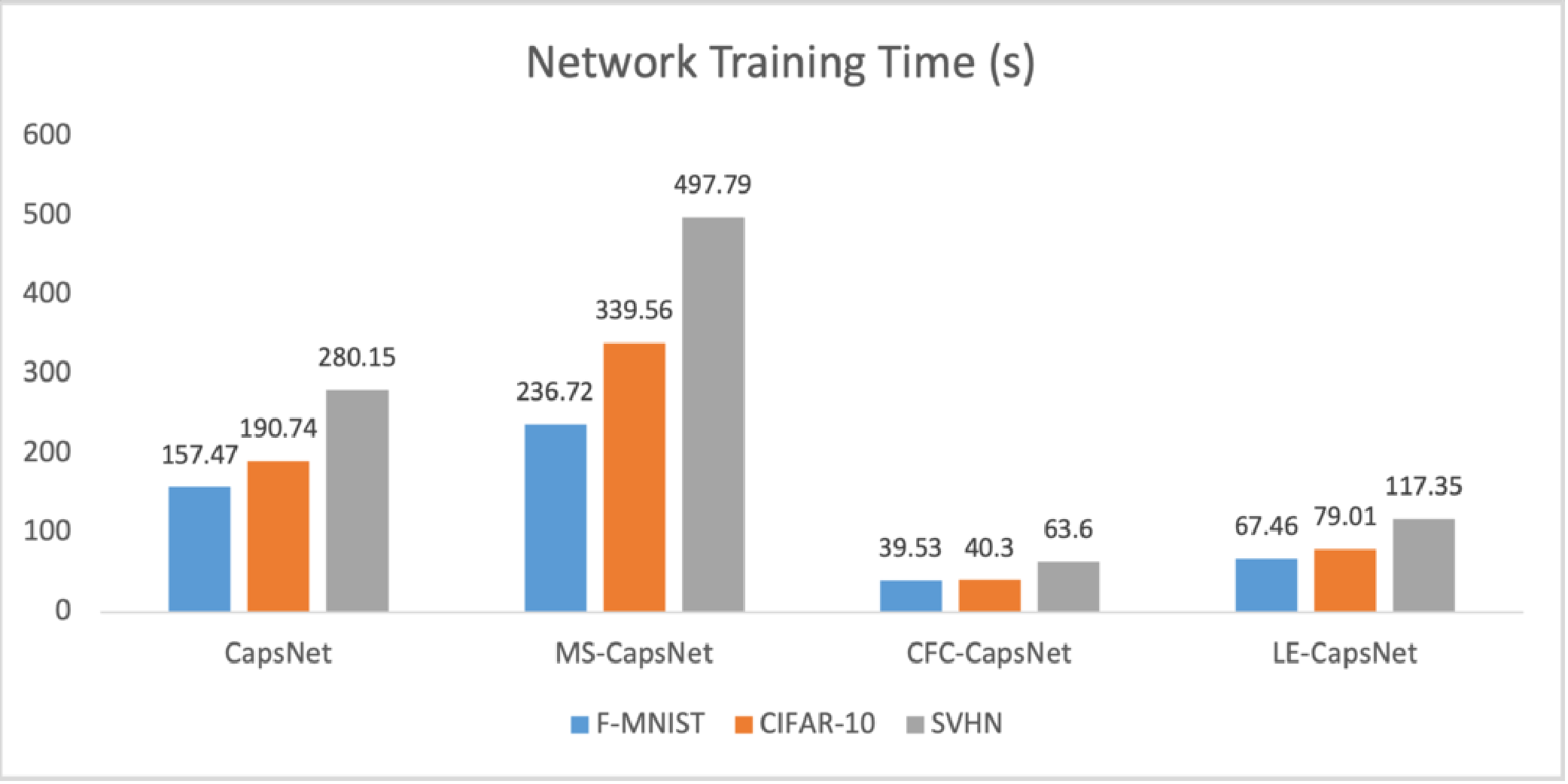}
    %[width=\textwidth,height=\textheight]
    \caption{Comparing the network training time between different architectures. LE-CapsNet is not as fast as CFC-CapsNet as it generates more capsules on three different scales.}
    \label{fig:res-train-speed}
\end{figure}
\begin{figure}[htp]
    \centering
    % \captionsetup{width=\textwidth,height=\textheight}
    \includegraphics[keepaspectratio, scale=0.2]{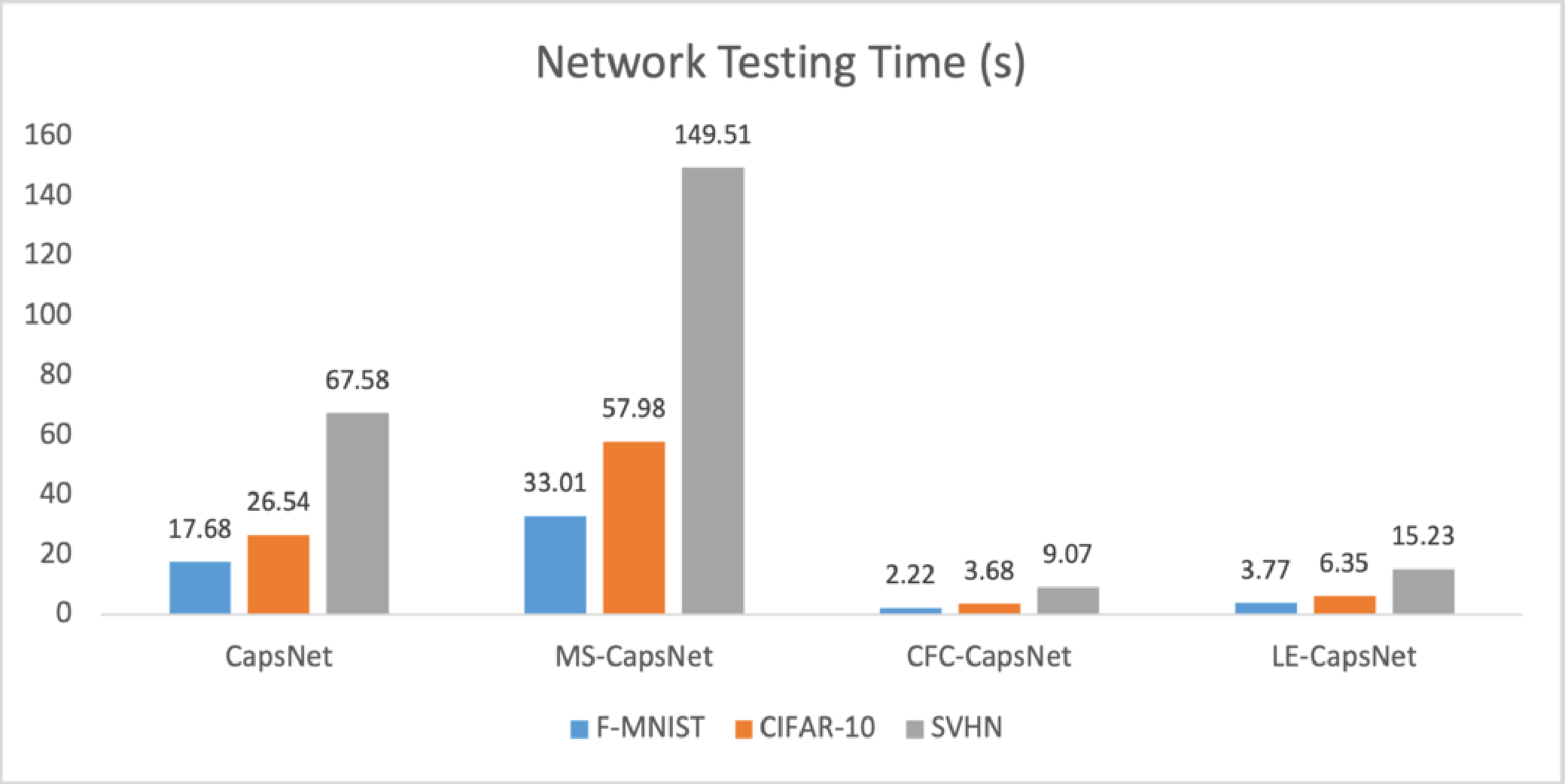}
    %[width=\textwidth,height=\textheight]
    \caption{Comparing the network testing time between different architectures. The results are consistent with the results from the training times: LE-CapsNet obtains the second highest speed.}
    \label{fig:res-test-speed}
\end{figure}

% We have access to the implementations of CapsNet, CFC-CapsNet and MS-CapsNet all in the PyTorch framework. So we used them to make a fair comparison between these networks and LE-CapsNets in terms of the training and inference times. 
Figure \ref{fig:res-train-speed} and \ref{fig:res-test-speed} reports the network training and testing times. LE-CapsNet is faster than CapsNet (~2.4x and ~4x faster training and testing respectively), but not as fast as CFC-CapsNet.This is due to the fact that there are 3 scales in the feature extractor and therefore LE-CapsNet has  more capsules compared to CFC-CapsNet.

\subsection{Robustness to Affine Transformations}
Sabour et al. \cite{Sabour2017} use the AffNIST dataset \footnote{https://www.cs.toronto.edu/~tijmen/affNIST/} to verify the robustness of CapsNet against applying an affine transformation to input images. This dataset includes 32 variations of the MNIST dataset each of which includes different types of affine transformations. We follow the same method. First, we create the "expanded MNIST" dataset which includes translations applied to MNIST images. We randomly place the 28x28 images of MNIST on a 40x40 grid to create the dataset. Once the network reaches 99.23\% of accuracy on the expanded MNIST dataset, it is tested against the AffNIST dataset. LE-CapsNet performs better than CapsNet: it achieves 94.37\% accuracy (compared to 90.52\% achieved by CapsNet) on the AffNIST dataset.

\par

\section{Conclusion}
We presented LE-CapsNet as a novel variant of CapsNet. The proposed network includes a specialized feature extractor that performs  in multiple scales. It also integrates a capsule translation module to produce capsules out of the extracted features. This implementation achieves a higher accuracy using fewer capsules and significantly fewer weights. For the CIFAR-10 dataset, LE-CapsNet improves accuracy from 71.69\% up to 76.73\% while reducing the number of weights by 67\%. The network is also 4 times faster during the inference.

\section*{Acknowledgment}
This research has been funded in part or completely by the Computing Hardware for Emerging Intelligent Sensory Applications (COHESA) project. COHESA is financed under the National Sciences and Engineering Research Council of Canada (NSERC) Strategic Networks grant number NETGP485577-15.

\bibliographystyle{unsrt}

\begin{thebibliography}{10}

\bibitem{Zhao2017a}
Bo~Zhao, Jiashi Feng, Xiao Wu, and Shuicheng Yan.
\newblock {A survey on deep learning-based fine-grained object classification and semantic segmentation}.
\newblock {\em International Journal of Automation and Computing}, 14(2):119--135, 2017.

\bibitem{Xiang2018}
Canqun Xiang, Lu~Zhang, Yi~Tang, Wenbin Zou, and Chen Xu.
\newblock {MS-CapsNet: A Novel Multi-Scale Capsule Network}.
\newblock {\em IEEE Signal Processing Letters}, 25(12):1850--1854, dec 2018.

\bibitem{Shiri2021}
P.~Shiri and A.~Baniasadi.
\newblock {Convolutional Fully-Connected Capsule Network (CFC-CapsNet)}.
\newblock In {\em ACM International Conference Proceeding Series}, 2021.

\bibitem{Sabour2017}
Sara Sabour, Nicholas Frosst, and Geoffrey~E Hinton.
\newblock {Dynamic Routing Between Capsules}.
\newblock (Nips), 2017.

\bibitem{shiri2020}
Pouya Shiri, Ramin Sharifi, and Amirali Baniasadi.
\newblock {Quick-CapsNet (QCN): A Fast Alternative to Capsule Networks}.
\newblock In {\em Proceedings of IEEE/ACS International Conference on Computer Systems and Applications, AICCSA}, volume 2020-Novem, 2020.

\bibitem{DoRosario2019}
Vanderson~Martins {Do Rosario}, Edson Borin, and Mauricio Breternitz.
\newblock {The Multi-Lane Capsule Network}.
\newblock {\em IEEE Signal Processing Letters}, 26(7):1006--1010, 2019.

\bibitem{Rajasegaran2019}
Jathushan Rajasegaran, Vinoj Jayasundara, Sandaru Jayasekara, Hirunima Jayasekara, Suranga Seneviratne, and Ranga Rodrigo.
\newblock {DeepCaps: Going Deeper with Capsule Networks}.
\newblock 2019.

\bibitem{Rajasegaran}
Jathushan Rajasegaran, Vinoj Jayasundara, Sandaru Jayasekara, Hirunima Jayasekara, Suranga Seneviratne, and Ranga Rodrigo.
\newblock {Deepcaps: Going deeper with capsule networks}.
\newblock In {\em Proceedings of the IEEE Computer Society Conference on Computer Vision and Pattern Recognition}, volume 2019-June, pages 10717--10725, 2019.

\bibitem{Xiao2017}
Han Xiao, Kashif Rasul, and Roland Vollgraf.
\newblock {Fashion-MNIST: a Novel Image Dataset for Benchmarking Machine Learning Algorithms}.
\newblock aug 2017.

\bibitem{Netzer2011}
Yuval Netzer, Tao Wang, Adam Coates, Alessandro Bissacco, Bo~Wu, and Andrew~Y. Ng.
\newblock {The Street View House Numbers (SVHN) Dataset}, 2011.

\bibitem{Krizhevsky2009}
A~Krizhevsky, V~Nair, and G~Hinton.
\newblock {CIFAR-10 and CIFAR-100 datasets}, 2009.

\end{thebibliography}

\end{document}